\pdfoutput=1

\documentclass[11pt]{article}

\usepackage[]{acl}

\usepackage{times}
\usepackage{latexsym}
\usepackage{algorithm}
\usepackage{algorithmic}
\usepackage{url}
\usepackage{multirow}
\usepackage{graphicx}
\usepackage{amssymb}
\usepackage{color}
\usepackage{caption}
\usepackage{subcaption}
\usepackage{amsmath}
\usepackage{graphicx}
\usepackage[T1]{fontenc}

\usepackage[utf8]{inputenc}
\usepackage{microtype}

\title{KD-VLP: Improving End-to-End Vision-and-Language Pretraining with \\Object Knowledge Distillation}

  \author{Yongfei Liu\textsuperscript{\rm 1, 4}\enspace Chenfei Wu\textsuperscript{\rm 3}\enspace Shao-yen Tseng\textsuperscript{\rm 2}\enspace Vasudev Lal\textsuperscript{\rm 2} \enspace Xuming He\textsuperscript{\rm 1} \enspace Nan Duan\textsuperscript{\rm 3}
  \\  \textsuperscript{\rm 1} ShanghaiTech \enspace \textsuperscript{\rm 2} Intel Labs, Cognitive Computing Research \enspace \textsuperscript{\rm 3} Microsoft Research Asia\\
  \textsuperscript{\rm 4} Shanghai Institute of Microsystem and Information Technology, Chinese Academy of Sciences\\
  \{liuyf3, hexm\}@shanghaitech.edu.cn\enspace \{chewu, nanduan\}@microsoft.com\\\{vasudev.lal, shao-yen.tseng\}@intel.com}

\begin{document}
\maketitle

\begin{abstract}
  Self-supervised vision-and-language pretraining (VLP) aims to learn transferable multi-modal
  representations from large-scale image-text data and to achieve strong performances on a broad scope of vision-language tasks
  after finetuning. 
  Previous mainstream VLP approaches typically adopt a two-step strategy relying on external object detectors to encode images in a multi-modal Transformer framework, which suffer from restrictive object concept space, limited image context and inefficient computation. In this paper,  
  we propose an object-aware end-to-end VLP framework, which directly feeds image grid features from CNNs into the Transformer
  and learns the multi-modal representations jointly. More importantly, we propose to perform object knowledge
  distillation to facilitate learning cross-modal alignment at different semantic levels. To achieve that, 
  we design two novel pretext tasks by taking object features and their semantic labels from external detectors as supervision: 
  1.) \textit{\textbf{Object-guided masked vision modeling}} task focuses on enforcing object-aware representation learning in the multi-modal Transformer;
  2.) \textit{\textbf{Phrase-region alignment}} task aims to improve cross-modal alignment by utilizing the similarities 
  between noun phrases and object labels in the linguistic space.
  Extensive experiments on a wide range of vision-language tasks demonstrate the efficacy of our proposed framework, and  
  we achieve competitive or superior performances over the existing pretraining strategies.
  
\end{abstract}

\section{Introduction}
With the success of BERT~\cite{devlin2018bert} in language modeling, self-supervised Vision-and-Language Pretraining (VLP)
has attracted much interest from AI community, which aims to learn generalizable multi-modal representations from large-scale image-text data. 
Combined with a pretrain-then-transfer strategy, it shows great potential in tackling vision and language reasoning tasks, such as image-text retrieval, 
visual question answering (VQA) and visual entailment~\cite{antol2015vqa,lee2018stacked,xie2019visual,Liu_2021_CVPR,liu2020learning}. 
A critical step in such representation learning is to jointly model linguistic entities and visual semantic concepts (e.g., attributes, objects, and relations), as well as their alignment. However, this is particularly challenging due to large discrepancy in visual and language representations (pixels vs words)  and lack of entity-level cross-modal correspondence in supervision.  

To tackle those challenges, most existing approaches~\cite{li2020unimo,gan2020large,chen2019uniter,lu2019vilbert} adopt 
a two-step pretraining strategy that firstly utilizes off-the-shelf detectors to parse images into a set of object tokens, and
then builds a multi-layer Transformer to learn visual and language embeddings jointly. 
In order to facilitate the multi-modal learning, those networks are typically trained via a set of carefully designed BERT-like objectives (e.g. Image-Text Matching). Despite its promising performance, the two-step strategy suffers from several limitations: 1) limited visual object concepts as the external detectors are trained on a predefined set of object categories; 2) lack of context cues outside of the object regions, which are crucial for complex reasoning tasks; 3) sub-optimal visual representation due to stage-wise training; and 4) computational inefficiency caused by additional detection modules. 
To overcome those limitations, recent works attempt to learn a joint visual-linguistic representations in an end-to-end manner~\cite{huang2021seeing,huang2020pixel,xu2021e2e,kim2021vilt}. These methods directly take dense visual features from image grids as inputs to a multi-modal Transformer network, and hence do not rely on external object detectors in both pretraining and finetuning stages. Such model design significantly simplifies overall network architecture and allows deeper integration between visual and language features. However, using grid-level features makes it difficult to capture object-level visual concepts, which often results in less expressive multi-modal representations and inferior performances in downstream tasks.   

In this work, we propose a novel object-aware end-to-end (E2E) VLP approach that inherits the strengths of both types of pretraining strategies mentioned above. Our core idea, which we name KD-VLP, 
is to incorporate visual object concepts in the E2E multi-modal learning, 
which is instantiated by performing \textbf{K}nowledge \textbf{D}istillation from semantic objects (e.g., from the off-the-shelf detectors) during the pretraining stage. 
This allows the network to better capture object representations and hence facilitates learning the alignment of linguistic entities and visual concepts. To achieve this, 
we introduce two novel pretext tasks to perform object knowledge distillation based on a CNN+Transformer architecture: an object-based masked vision modeling task for enforcing object-aware feature embeddings, and a phrase-region alignment task for  
building correspondence between object regions and language entities. 

Specifically, we adopt a typical CNN backbone+multi-modal Transformer model for the pretraining. Given an image-text pair, the visual backbone firstly computes a set of visual features on the image grid. Then a multi-layer Transformer takes the visual features and the corresponding text tokens as input to generate their multi-modal embeddings. Based on those embeddings, a set of task-specific heads compute the corresponding objectives to train the entire network in an end-to-end fashion. Here, in addition to the commonly-used image-text matching and masked language modeling objectives, we develop two object-aware pretext tasks. The first task, \textit{object-guided masked vision modeling} (OMVM), aims to reconstruct the RoI features and semantic label of each object (from an external detector) using the surrounding visual context and text description. To facilitate cross-modal alignment, we also develop a knowledge-guided masking strategy, which samples object candidates for reconstruction according to the similarity scores between the noun phrases in the corresponding text and their semantic labels.
The second task, \textit{phrase-region alignment} (PRA), aims to further improve cross-modal alignment by matching the above-mentioned phrase-label similarity scores of each phrase with the cross-modal similarity scores between the noun phrase embeddings and object region embeddings. 
After pretraining, we then transfer the learned multi-modal representations to different downstream vision-language tasks.
 
We perform pretraining on two widely-used indomain datasets: MSCOCO Caption~\cite{lin2014microsoft} and Visual Genome~\cite{krishna2016visual}, 
and validate the learned multi-modal representations on five well-known visual-language tasks: Visual Question Answering~(VQA), Image-text retrieval, 
Nature Language Visual Reasoning~(NLVR$^2$), Visual Entailment~(VE) and Visual Commonsense Reasoning~(VCR). Empirical results show that our method outperforms the state-of-the-art end-to-end approaches by a sizeable
margin. To better understand our method, we also provide a detailed ablation study and visualization.

The contributions of our work are three-fold:
\begin{itemize}
    \item We propose a novel end-to-end pretraining strategy,  capable of better encoding visual object concepts and 
    facilitating multi-modal representation learning.

    \item We design an object-guided masked vision model task for distilling knowledge from external object detectors, and 
    a phrase-region alignment task to facilitate learning better phrase-region correspondence.

    \item Compared with existing methods, we achieve competitive or superior performances without using
    external detection outputs during finetuning stage and model test.
    
\end{itemize}

\section{Related Work}
The existing self-supervised VLP approaches can be largely categorized into two groups: the two-step
pretraining and the end-to-end pretraining, depending on whether they rely on visual object embeddings as input for the Transformer.

\textbf{Two-step Pretraining} firstly employ an off-the-shelf object detector
to convert an image into a set of object embeddings, and then feed them into a Transformer jointly with text embeddings to generate their
multi-modal representations. Hence their visual feature networks are not optimized during both pretraining \& finetuning stage. 
Most of these methods, such as LXMERT~\cite{tan2019lxmert},ViLBert~\cite{lu2019vilbert}, VL-Bert~\cite{su2019vl}, Unicoder-VL~\cite{li2020unicoder} and 
UNITER~\cite{chen2019uniter}, adopt BERT-like objectives to 
train their networks, which include Masked Language Modeling (MLM), Masked Vision Modeling (MVM) and Image-Text Matching (ITM). 
In addition, VILLA~\cite{gan2020large} develops an advanced adversarial pretraining and finetuning strategy to improve generalization ability. 
OSCAR~\cite{li2020oscar} and VINVL~\cite{Zhang_2021_CVPR} introduce object labels to bridge different modalities and revisit the importance of visual features.
Ernie-ViL~\cite{yu2020ernie} exploits structured knowledge in the text and constructs scene graph prediction tasks to learn joint representations.
UNIMO~\cite{li2020unimo} proposes a unified model to leverage large-scale free text corpus, image collections, and image-text pairs simultaneously
through a contrastive learning task. 
Despite their strong performances, those methods are limited by the object detector and neglect visual cues outside of object regions, 
often leading to mistakes in downstream tasks.

\textbf{End-to-End (E2E) Pretraining} directly feed dense features on image grids from a visual backbone network into a Transformer network along with text tokens. As such, both the visual and Transformer networks are optimized jointly in an end-to-end manner in the pretraining \& finetuning stage.
Pixel-Bert and SOHO~\cite{huang2021seeing,huang2020pixel} pioneer the use of the E2E pretraining architecture and propose a novel visual-dictionary masked vision modeling task. E2E-VLP~\cite{xu2021e2e}
presents a pretraining framework supervised with additional object detection and image captioning tasks to enhance visual semantics learning. 
It is worth noting that their object detection pretext task requires millions of bounding boxes annotation, unable to generalize to large-scale image-text corpus.
ViLT~\cite{kim2021vilt} is the first to unify vision and language with a pure Transformer network, which has a simpler structure and enjoys faster inference.
However, compared to the two-step methods, they are typically less expressive in terms of object-level concepts and thus suffer from weaker performances on challenging visual reasoning tasks. Our method is in line with the E2E pretraining framework. The key difference is that we propose to facilitate learning object-aware multi-modal representations by performing object semantic knowledge distillation.

\section{Our Approach}
\begin{figure*}
    \centering
    \includegraphics[width=0.9\linewidth]{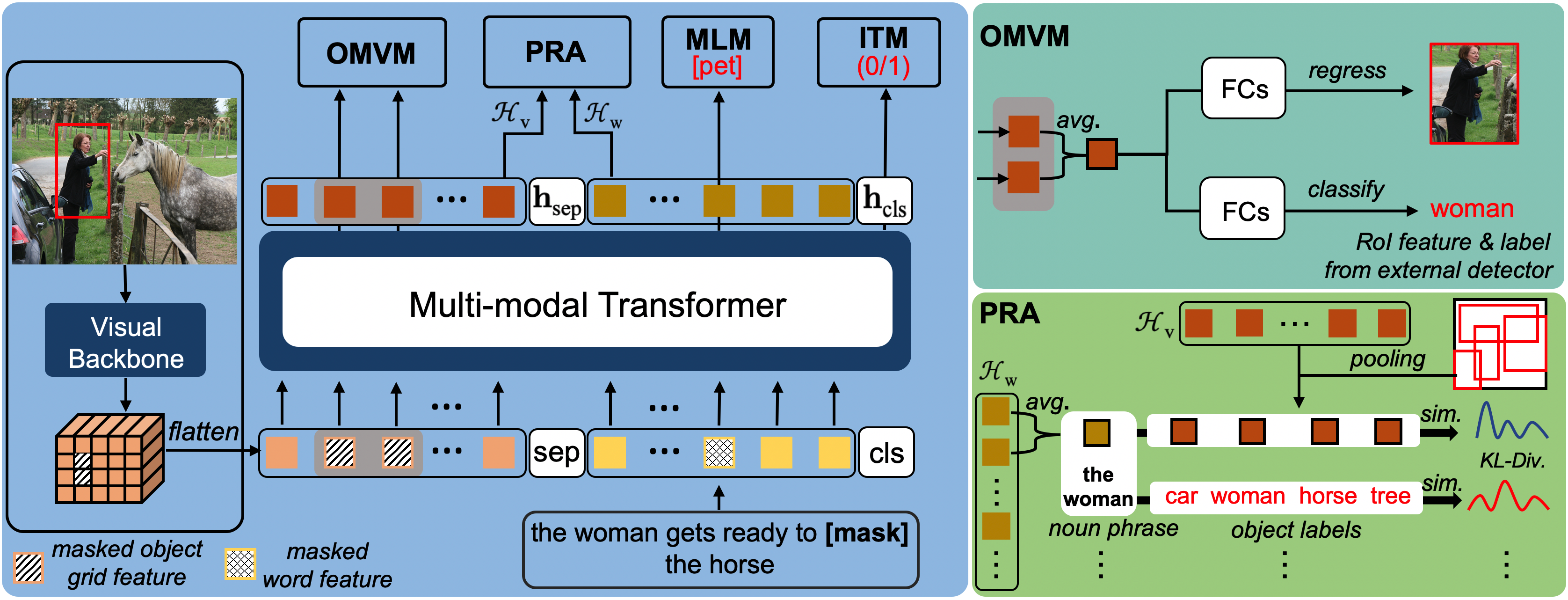}
    \caption{\small{\textbf{Overview:} The model contains a Visual Backbone for preparing image embeddings and
    a Transformer for vision \& language fusion. The entire framework is supervised by two novel proposed pretext tasks: 
    Object-guided Masked Vision Modeling (OMVM), Phrase-Region Alignment (PRA) as well as two standard tasks: Masked Language Modeling (MLM)
    and Image-Text Matching (ITM).}}
    \label{overview}
\end{figure*}

\subsection{Problem Definition and Overview}
The goal of self-supervised VLP is to learn a generic and transferable visual-linguistic representation from a large amount of image-text data, which can achieve strong generalization performances in downstream vision-language
tasks. To this end, the pretraining framework typically develops a variety of carefully-designed cross-modal pretext tasks (e.g. MLM, ITM) to train a deep network that encodes the multi-modal representation.
Formally, we denote the image-text corpus for training as $\mathcal{X}$ = $\{(I^i, D^i)\}^{|\mathcal{X}|}_{i=1}$ where $I$ represents the image and $D$ is the corresponding language description.
In general, we construct a pretraining network consisting of a representation network module $\mathcal{M}_\theta$ and a set of task-specific network heads $\{\Phi_{\theta_s}\}_{s=1}^S$ where $s$ indicates the pretext tasks.
The overall pretraining objective is defined as follows,
\begin{equation}
    \resizebox{.8\hsize}{!}{$\mathop{{\rm min}}\limits_{\theta, \theta_1,...\theta_S}  \mathbb{E}_{(I,D)\sim \mathcal{X}}[\sum_{s} L_s(Y_s, \Phi_{\theta_s}\circ \mathcal{M}_\theta(I, D)]$}
\end{equation}
where $Y_s$ and $L_s$ are task-specific ground-truth label and loss function respectively, and $\circ$ is a network compound operator. After pretraining, we remove all the task-specific heads and apply the representation network $\mathcal{M}_{\theta^*}$ with the learned parameters ${\theta^*}$ to the downstream tasks, followed by task-specific fine-tuning.

In this work, we aim to design an E2E pretraining strategy for the VLP problem. To this end, we adopt a modular representation network, which takes image grid features from a CNN-based visual network and the corresponding text embeddings into a multi-modal  
Transformer~\cite{huang2020pixel,huang2021seeing}. Our goal is to learn the visual network and the Transformer jointly, and yet to effectively encode object-level visual concepts in the multi-modal representations. 
This enables us to capture rich cross-modal alignment between linguistic entities and visual semantic concepts for the downstream tasks, 
and meanwhile to enjoy the benefits of an efficient E2E network design without relying on detectors during fine-tuning and inference.  

To achieve this, we propose a set of cross-modal pretext tasks that perform object knowledge distillation from external detectors in both semantic and feature space. 
Specifically, in addition to the image-text matching (ITM) and masked language modeling (MLM) tasks, we introduce two novel pretext tasks, 
Object-Guided Masked Vision Modeling (OMVM) and Phrase-Region Alignment (PRA), which take the object RoI feature embeddings and semantic labels from external detectors as supervision. The OMVM task masks out the object regions and forces the network to predict the corresponding external RoI feature embeddings and object labels while the PRA task exploits object labels to encourage the alignment between visual objects and language entities. Fig.\ref{overview} illustrates an overview of our framework. 
Below we will first present the details of model architecture in Sec.\ref{sec:backbone}, followed by our design of pretext tasks in Sec.\ref{sec:tasks}.

\subsection{Model Architecture}
Given an image-text pair, our model firstly computes the image embeddings and linguistic embeddings respectively, and then concatenates them into
a sequence of tokens with two additional tokens \textbf{[sep]} and \textbf{[cls]} as inputs to a Transformer for generating multi-modal contextualized embeddings.

\paragraph{Visual Embedding} We adopt a CNN backbone to extract image features $\mathcal{V}$ = $\{\mathbf{v}_i\}_{i=1}^L$ for 
each image $I$ where $L$ is the size of feature grids and $\mathbf{v}_i\in \mathcal{R}^{d_v}$ is a
feature vector with dimension $d_v$. In addition, each feature is further concatenated with its
2-D sine position embedding~\cite{carion2020end}. Following SOHO, we use a ResNet-101\cite{He_2016_CVPR} as the visual backbone, followed by additional 1x1 Conv and 2x2 strides Max-pooling to reduce the memory footprint.

\paragraph{Linguistic Embedding} 
For the language $D$, we first tokenize the sentence into a sequence of word tokens using WordPiece~\cite{wu2016google},
then encode them into word embeddings $\mathcal{W}$ = $\{\mathbf{w}_j\}_{j=1}^T$ where $\mathbf{w}_j\in \mathcal{R}^{d_w}$ is the feature vector. Similarly, an index position~\cite{devlin2018bert} 
embedding is supplemented to each word embedding.

\paragraph{Multi-modal Transformer}
After obtaining image and linguistic embeddings, we assemble them into a sequence of tokens $\{\mathcal{V}, \mathbf{[sep]},\mathcal{W},\mathbf{[cls]}\}$, 
and adopt a multi-layer Transformer to compute their representations encoded by the final-layer 
states $\{\mathcal{H}_{\mathcal{V}}, \mathbf{h}_{\mathbf{sep}}, \mathcal{H}_{\mathcal{W}}, \mathbf{h}_{\mathbf{cls}}\}$ 
where $\mathcal{H}_{\mathcal{V}}$ = $\{\mathbf{h}_{\mathbf{v}_i}\}_{i=1}^L$  and 
$\mathcal{H}_{\mathcal{W}}$ = $\{\mathbf{h}_{\mathbf{w}_j}\}_{j=1}^T$ represent the states for visual and language part respectively.
Finally, those representations are sent into each pretext task head to compute the supervision signals.

\label{sec:backbone}
\subsection{Pretext Tasks} 
\label{sec:tasks}
We now describe our cross-modal pretext tasks for the E2E pretraining, 
aiming to learn more effective multi-modal representations. 
Below we will first introduce objects-aware pretext tasks that take external 
object features and semantic labels as supervision, followed by the standard MLM and ITM. 

Specifically, for each image, we first generate a set of object proposals from an off-the-shelf detector, denoted as
$\{(o_n, c_n, \mathbf{f}_n)\}_{n=1}^N$ where $o_n\in \mathcal{R}^4$ is box location, $c_n$ indicates object category, and
$\mathbf{f}_{n}\in \mathcal{R}^{d_o}$ is object RoI features with dimension $\mathcal{R}^{d_o}$. For ease of notation, we also introduce a binary mask\footnote{We give an illustration in Suppl.} on the feature map for each object $o_n$ and denote its flattened version as  $\mathbf{m}_{n}\in \mathcal{R}^L$. 
For the corresponding text, we extract a set of 
noun phrases $\mathcal{P}$ = $\{p_z\}_{z=1}^{|\mathcal{P}|}$ with an external language tool\footnote{https://spacy.io/} and calculate the similarity
$\alpha_{z,n}$ between each noun phrase $p_z$ and the object category $c_n$ in the linguistic space:

\begin{equation}
    \resizebox{0.6\hsize}{!}{$\alpha_{z,n} = {\rm Cos}(E_{ext}(p_z), E_{ext}(c_n)),$} 
\end{equation}
where ${\rm Cos}(\cdot,\cdot)$ indicates cosine distance function and $E_{ext}$ represents an off-the-shelf language embedding (e.g. BERT). 
Using them as supervision, we design two novel pretext tasks to distill object-level knowledge below.

\paragraph{Object-guided Masked Vision Modeling (OMVM)} 
The first task aims to learn more explicit object concepts in the E2E pretraining. Specifically, we sample an object each time and mask out its features in the Transformer input, and enforce the network to generate external object RoI features and semantic labels.  
To learn better cross-modal alignment, we propose a knowledge-guided masking strategy, which 
samples noun phrase-related object regions to mask based on the (normalized) similarity score $\alpha_{z,n}$.
The selected object region is denoted with its binary mask, category and RoI features, as $(\mathbf{m}^*, c^*, \mathbf{f}^*)$.

We design two learning objectives, Masked Region Classification (MRC) and Masked Region Feature Regression (MRFR) as below
\begin{equation}
    \begin{aligned}
        \resizebox{0.17\hsize}{!}{$\mathcal{L}_{{\rm OMVM}} =$} & \resizebox{0.5\hsize}{!}{$\mathbb{E}_{(I,D)\sim \mathcal{X}} L_{{\rm MRC}}(c^*, \mathcal{V}_{\backslash \mathbf{m}^*}, \mathcal{W})$} \\
                                   & \resizebox{0.4\hsize}{!}{$+ L_{{\rm MRFR}}(\mathbf{f}^*, \mathcal{V}_{\backslash \mathbf{m}^*}, \mathcal{W})$}
    \end{aligned}
\end{equation}
To calculate the losses $L_{\rm MRC}$ and $L_{\rm MRFR}$, we first compute the object representation $\mathbf{h}_{\mathbf{m}^*}$ for the masked region at the final layer, 
which is average-pooled over $\mathcal{H}_{\mathcal{V}}$ based on its binary mask $\mathbf {m}^*$. 
For MRC, a multi-layer FC network $\Phi_{{\rm MRC}}$ is adopted to predict its object category. 
Thus, $L_{{\rm MRC}}$= ${\rm CE}(\Phi_{\rm{MRC}}(\mathbf{h}_{\mathbf{m}^*}), c^*)$ is the standard cross-entropy loss.
In addition, we take another FC network $\Phi_{{\rm MRFR}}$ to learn the object concept in feature space directly by minimizing
the L2 distance, $L_{{\rm MRFR}}$ = $||\Phi_{{\rm MRFR}}(\mathbf{h}_{\mathbf{m}^*}) - \mathbf{f}^*)||_2^2$.

\paragraph{Phrase Region Alignment (PRA)}
The second task, PRA, mainly focuses on learning cross-modal alignment at object-level, which aims to pull positive phrase-region pairs closer and push negative pairs away.
Here we utilize the similarity $\alpha_{z,n}$ between the noun phrase and object category in the linguistic space as a guidance.

Concretely, we first compute the object representation ${\mathbf{h}_{\mathbf{m}_n}}$ for each proposal and
the phrase representation $\mathbf{h}_{p_z}$, both of which are obtained from the final layer states of the Transformer. 
Specifically, ${\mathbf{h}_{\mathbf{m}_n}}$ is average-pooled over $\mathcal{H}_{\mathcal{V}}$ based on binary mask ${\mathbf{m}}_n$
while $\mathbf{h}_{p_z}$ = $\frac{1}{|p_z|} \sum_{j\in p_z}$ $\mathbf{h}_{\mathbf{w}_j}$ represents average states of word tokens within $p_z$.
We define the cross-modal similarity as $\hat{\alpha}_{z,n}$ = ${\rm Cos}(\mathbf{h}_{p_z}, {\mathbf{h}_{\mathbf{m}_n}})$. 

The task PRA minimizes the KL-divergence between the cross-modal similarities $\hat{\alpha}_z$ = $\{{\rm Softmax}(\hat{\alpha}_{z,n})\}_{n=1}^N$ and the phrase-label similarities $\alpha_z$ = $\{{\rm Softmax}(\alpha_{z,n})\}_{n=1}^N$ as below: 

\begin{equation}
	\resizebox{0.58\hsize}{!}{$L_{{\rm PRA}} =  \frac{1}{|\mathcal{P}|} \sum_z D_{KL} (\hat{\alpha}_z || \alpha_z)$}
\end{equation}
Finally, denoting the mask set $\mathcal{M}=\{\mathbf{m}_n\}_{n=1}^N$, we have the overall PRA loss function as follows:
\begin{equation}
    \resizebox{0.88\hsize}{!}{$\mathcal{L}_{{\rm PRA}} =  \mathbb{E}_{(I, D)\sim \mathcal{X}} L_{{\rm PRA}}( \{\alpha_{z,n}\}_{z,n=1}^{|\mathcal{P}|, N}, \mathcal{M},\mathcal{P}, \mathcal{V}, \mathcal{W})$}
\end{equation}

\paragraph{Masked Language Modeling (MLM)} We take the same masking strategy (15\% \textit{prob.} to mask) as 
in BERT~\cite{devlin2018bert} to randomly mask out the input word tokens. Here, 
MLM aims to predict the original word index in vocabulary space for each masked 
token based on the whole image and its surrounding language context via the Transformer. Hence a cross-entropy loss is adopted:

\begin{equation}
    \resizebox{0.7\hsize}{!}{$\mathcal{L}_{{\rm MLM}} = -\mathbb{E}_{(I, D)\sim \mathcal{X}} {\rm log} P(\mathbf{w}_j|\mathcal{V}, \mathcal{W}_{\backslash j})$}
\end{equation}

\paragraph{Image-Text Matching (ITM)} In ITM, the multi-layer Transformer is trained to distinguish whether the input image-text pairs are semantically matched based on the 
final layer \textbf{[cls]} token representation $\mathbf{h}_{\mathbf{cls}}$. To construct the training samples, we randomly replace the text for each image-text pair with another text from dataset with a probability of 0.5. Thus, the output
label can be defined as $y\in \{0, 1\}$ where $y=1$ indicates matched pair. The training objective for the ITM task is to minimize binary cross-entropy loss:

\begin{equation}
    \resizebox{0.65\hsize}{!}{$\mathcal{L}_{{\rm ITM}} = -\mathbb{E}_{(I, D)\sim \mathcal{X}} {\rm log} P(y|\mathcal{V}, \mathcal{W})$}
\end{equation}

\begin{table*}
	\centering
	\caption{\small{Evaluation results on the multi-modal downstream tasks. \textbf{Indomain} 
    denotes MSCOCO and Visual Genome corpus while \textbf{outdomain} stands for Conceptual Caption and SBU corpus. \textbf{Text corpus} includes BookWiki and OpenWebText while \textbf{image corpus} contains OpenImages and unlabeled COCO.
    AT means using adversarial training during both pretraining and finetuning stages. \textcolor{blue}{Blue number} denotes experiments with additional text premise input. - denotes the result is not available}}
    \resizebox{1.0\textwidth}{!}{
		\begin{tabular}{ccccccccc}
			\hline
			\multirow{2}{*}{Models}         &\multirow{2}{*}{Pretraining corpus} &\multirow{2}{*}{Backbone} &\multirow{2}{*}{AT}       &Flickr30k-IR          &Flickr30k-TR   &SNLI-VE    &NLVR$^2$     &VQA2.0                  \\
			                                &                         &    &                                      &R@1~/~R@5/~R@10          &R@1~/~R@5~/~R@10   &val~/~test   &dev~/~test-p      &test-dev~/~-std           \\
			\hline
            \textit{\textbf{two-step pretraining}}\\
            ViLBert~\cite{lu2019vilbert}      &Conceptual Cap.    &ResNet101        &x  &58.20~/~84.90~/~91.52     & -             & -         &-         &70.55~/~70.92             \\
            VL-Bert~\cite{su2019vl}           &Conceptual Cap.    &ResNet101        &x        &-&-&-&-&71.79~/~72.91\\
            VisualBert~\cite{li2019visualbert}&MSCOCO             &ResNet152        &x  &71.33~/~84.98~/~86.51&-&-&67.40~/~67.00&70.80~/~71.00\\
            Unicoder-VL\cite{li2020unicoder}  &outdomain          &ResNet101        &x  &71.50~/~90.90~/~94.90  &86.20~/~96.30~/99.00&-&-&-\\
            LXMERT~(Tan et al. 2019)          &indomain           &ResNet101        &x  &-&-&-&74.90~/~74.50 &72.42~/~72.54\\
            VLP~\cite{zhou2020unified}        &outdomain          &ResNext101       &x  &-     & -             & -         &-         &70.50~/~70.70             \\
            UNITER~\cite{chen2019uniter}      &indomain+outdomain &ResNet101        &x  &72.52~/~92.36~/~96.08     &85.90~/~97.10~/~98.80    &78.59~/~78.28     &75.85/75.80  &72.70~/~72.91     \\
            OSCAR~\cite{li2020oscar}          &indomain+outdomain &ResNet101        &x  &-      &-&-&78.07~/~78.36&72.16~/~73.44      \\
            VILLA~\cite{gan2020large}         &indomain+outdomain &ResNet101        &\checkmark   &74.74~/~92.86~/~95.82     &86.60~/~97.70~/~99.20   &79.47~/~79.03 &78.39~/~79.30 &73.59~/~73.67 \\
            Ernie-ViL~\cite{yu2020ernie}      &outdomain& ResNet101                 &\checkmark   &74.44~/~92.72~/~95.94     &86.70~/~97.80~/~99.00  &-      &- & 72.62~/~72.85 \\
            \multirow{2}{*}{UNIMO~\cite{li2020unimo}}         &indomain+outdomain+& \multirow{2}{*}{ResNet101}     &\multirow{2}{*}{\checkmark}  &\multirow{2}{*}{74.66~/~93.40~/~96.08}     &\multirow{2}{*}{89.70~/~98.40~/~99.10}  &\multirow{2}{*}{80.00~/~79.10} &\multirow{2}{*}{-} & \multirow{2}{*}{73.79~/~74.02} \\
                                             &text-corpus+ image-corpus               & & &&&&&\\
            \hline 
            \textit{\textbf{end-to-end pretraining}} \\
            Pixel-Bert~\cite{huang2020pixel}  &indomain           &ResNet50           &x  &59.80~/~85.50~/~91.60         &87.00~/~98.90~/~99.50  &-&71.70~/~72.40 &71.35~/~71.42 \\
            E2E-VLP ~\cite{xu2021e2e}         &indomain           &ResNet101            &x  &-&-&-&75.23~/-~&72.43~/~- \\
            ViLT~\cite{kim2021vilt}          &indomain+outdomain& ViT-B      &x  &64.40~/~88.70~/~93.80     &83.50~/~96.70~/~98.60  &- &75.70~/~76.13  &71.26~/~-\\
            SOHO ~\cite{huang2021seeing}      &indomain          & ResNet101  &x  &72.50~/~92.70~/~96.10     &86.50~/~98.10~/~99.30  &\textcolor{blue}{85.00}~/~\textcolor{blue}{84.95}  &76.37~/~77.32 & 73.25~/~73.47 \\
            KD-VLP~(ours)                       &indomain          & ResNet101  &x  &78.20~/~94.56~/~97.02     &91.40~/~98.90~/~99.40  &78.21(\textcolor{blue}{88.18})~/~77.87(\textcolor{blue}{88.21}) &77.36~/~77.78  &74.20~/~74.31\\
			\hline
	\end{tabular}}
	\label{res_tab1}
\end{table*}
\begin{table*}
	\centering
	\caption{\small{ Evaluation of image retrieval (IR) and text retrieval (TR) task on MSCOCO dataset and the performance of VCR task.}}
	\resizebox{0.8\textwidth}{!}{
		\begin{tabular}{ccccccccc}
			\hline
			\multirow{2}{*}{Models}         &\multirow{2}{*}{Backbone}   &MSCOCO-IR(1K)      &MSCOCO-TR(1K)      &MSCOCO-IR(5K)      &MSCOCO-TR(5K)           &\multicolumn{3}{c}{VCR} \\
                                            &                            &R@1~/~R@5~/~R@10   &R@1~/~R@5~/~R@10   &R@1~/~R@5~/~R@10   &R@1~/~R@5~/~R@10        & Q$\rightarrow$A &QA$\rightarrow$R & Q$\rightarrow$AR\\
			\hline
            \textit{\textbf{two-step pretraining}} \\
            Unicoder-VL\cite{li2020unicoder}  & ResNet101         &69.70~/~93.50~/~97.20  &84.30~/~97.30~/~99.30 &46.70~/~76.00~/~85.30 &62.30~/~87.10~/~92.80  &72.60  &74.50   &54.40\\
            UNITER~\cite{chen2019uniter}      & ResNet101         &-&-                                     &50.30~/~78.50~/~87.20 &64.40~/~87.40~/~93.10        &74.56  &77.03   &57.76\\
            OSCAR~\cite{li2020oscar}          & ResNet101         &-&-                                     &54.00~/~80.80~/~88.50 &70.00~/~91.10~/~95.50        &-  &-   &-\\
            VILLA~\cite{gan2020large}         & ResNet101         &-&-                                     &- &-         &75.54  &78.78   &59.75\\
            VL-Bert~\cite{su2019vl}         & ResNet101         &-&-                                     &- &-           &73.80  &74.40   &55.20\\
            \hline 
            \textit{\textbf{end-to-end pretraining}} \\
            Pixel-Bert~\cite{huang2020pixel}  & ResNet50         &64.10~/~91.00~/96.20 &77.80~/95.40~/~98.20   &41.10~/~/69.70~/~80.50  &53.40~/~80.40~/~88.50 &-&-&-             \\
            ViLT~\cite{kim2021vilt}          & ViT-B            &-&- &42.70~/~72.90~/~83.10 &61.50~/~86.30~/~92.70 &-  &-   &-\\
            SOHO~\cite{huang2021seeing}       & ResNet101        &73.50~/~94.50~/~97.50  &85.10~/~97.40~/~99.40  &50.60~/~78.00~/~86.70 &66.40~/~88.20~/~93.80 &-  &-   &- \\
            KD-VLP~(ours)                       & ResNet101        &75.21~/~94.89~/~97.99&88.62~/~98.18~/~99.44  &56.64~/~82.17~/~89.49 &74.28~/~92.86~/~96.28 &76.70  &78.63   &60.54 \\
			\hline
	 \end{tabular}}
	\label{res_tab2}
\end{table*}

\section{Experiments}
\subsection{Experiment Setup}

\paragraph{Pretraining Corpus:} Following the E2E pretraining strategy~\cite{huang2021seeing,huang2020pixel,xu2021e2e}, 
we take indomain datasets: MSCOCO ~\cite{lin2014microsoft} and VG~\cite{krishna2016visual} as pretraining datasets since it is widely used in literature.
In total, two datasets comprise about 200K images and 5.6M image-text pairs, where each image is associated with multiple captions. 

\paragraph{Implementation Details:} We follow BERT to tokenize caption into word tokens by using WordPiece,
and resize the image into (800, 1333) as prior works.
For model architecture, a widely-used ResNet101 for visual encoding and 12-layer Transformer for multi-modal fusion are adopted for a fair comparison.
Both networks are initialized with ImageNet and BERT pretrained parameters. Besides, following the majority of two-step methods,
we apply the widely-used object detector BUTD~\cite{anderson2018bottom} to generate object proposals as well as their RoI embeddings as our supervision.

For model learning, we optimize the entire network by using SGD for CNNs with a learning rate of 1e-2 and 
AdamW for Transformer with a learning rate of 1e-4, as suggested in SOHO. 
The training iterations are up to 100K with batch-size 512 in each. The learning rate decays 10 times at 20K, 40K respectively. 
All experiments are conducted on 16 NVIDIA V100 GPUs with mixed-precision training to reduce memory cost about 7 days. 

\vspace{-2mm}
\subsection{Downstream Tasks}
As in prior works, we evaluate our approach by finetuning it over a set of well-established VL understanding tasks, including
image-text retrieval, visual entailment (VE), natural language visual reasoning (NLVR$^2$), VQA, and VCR.
During finetuning, we compound a specific learnable head with the pretrained visual backbone and Transformer, then finetune the entire network with downstream task-specific
loss in an E2E fashion. In this work, we mainly compare performance with SOHO, Pixel-Bert, E2E-VLP, and ViLT since they are the E2E pretraining as ours.
Besides, several representative two-step pretraining approaches are also selected to compare without loss of generality.
Next, we will depict results analysis for each task and leave finetuning experiment setups in Suppl.

\begin{table*}
	\centering
    \caption{\small{Ablation study of various proposed pretext tasks. Image-text Retrieval task is conducted on MSCOCO 1K test set.}}
    \vspace{-2mm}
	\resizebox{0.8\textwidth}{!}{
		\begin{tabular}{ccccccc}
			\hline
			\multirow{2}{*}{Models}   &\multirow{2}{*}{Pretext Tasks}  &MSCOCO-TR(1K)          &MSCOCO-IR(1K)   &SNLI-VE    &NLVR$^2$     &VQA2.0                  \\
			                          &                           &R@1~/~R@5~/~R@10          &R@1~/~R@5~/~R@10   &val~/~test   &dev~/~test-p      &test-dev~/~-std \\
            \hline 
            baseline                  & ITM+MLM                 &57.99~/~87.80~/~94.66  & 73.10~/~93.42~/~97.32     &73.44~/~73.40 &62.13~/~62.08 &66.62~/~66.68 \\
            -                         & ITM+MLM+StandardMVM         &58.22~/~87.59~/~94.60  & 73.58~/~93.66~/~97.63     &74.00~/~73.46 &63.26~/~62.75 &66.66~/~66.86 \\
            -                         & ITM+MLM+RandomMVM       &58.18~/~87.12~/~94.68  & 73.60~/~94.80~/~97.50     &73.99~/~74.58 &64.02~/~64.68 &66.90~/~66.05 \\
            -                         & ITM+MLM+OMVM           &60.32~/~88.65~/~95.15  &74.83~/~94.34~/~97.74      &74.54~/~75.12 &66.23~/~66.76 &67.95~/~68.21 \\
            KD-VLP~(ours)               & ITM+MLM+OMVM+PRA       &61.10~/~89.40~/~95.50  &76.70~/~95.00~/~98.00      &74.62~/~75.22 &66.71~/~67.59 &68.19~/~68.43 \\
            \hline
	\end{tabular}}
	\label{aba}
\end{table*}
\begin{figure*}
    \centering
    \includegraphics[width=0.85\linewidth]{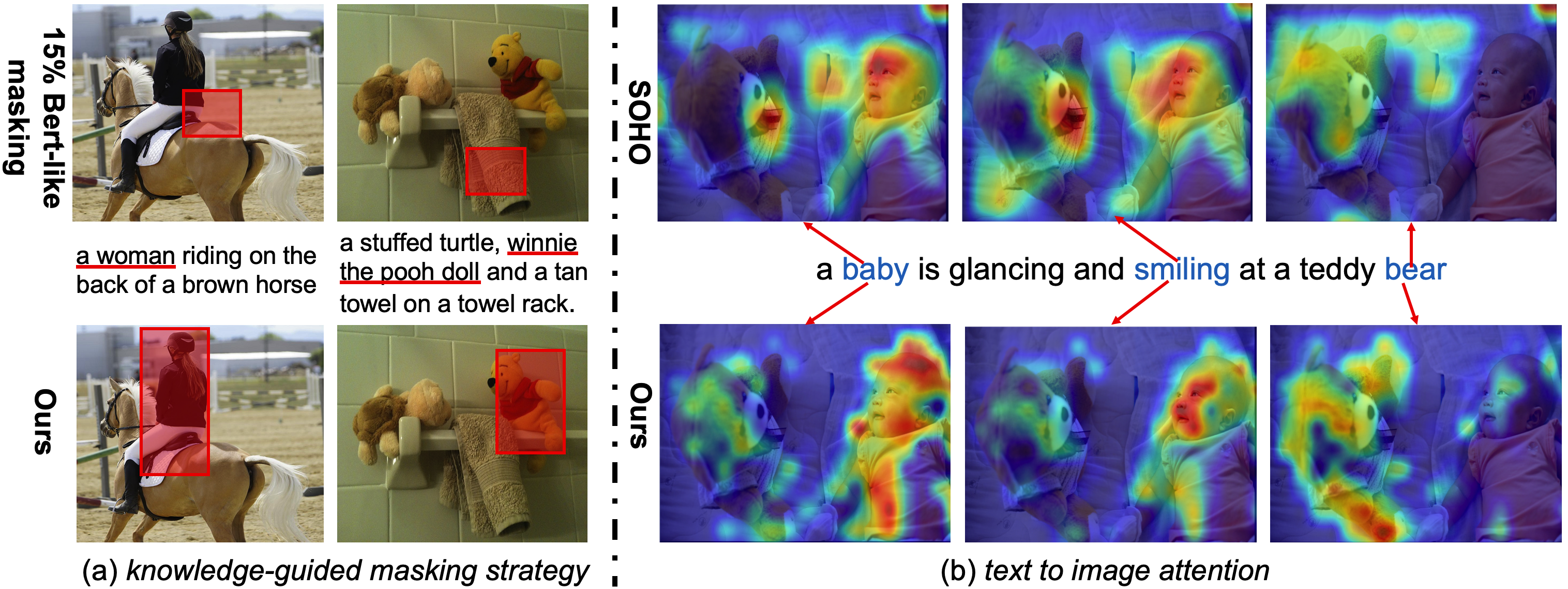}
    \caption{\small{(a) demonstrates the comparison of different masking vision strategies where the first row
    presents the 15\% Bert-like masking strategy adopted by all previous works and the second row shows our knowledge-guided masking strategy. \textcolor{red}{Red masks} denotes the masked regions.
    (b) demonstrates a comparison of word-to-image attention maps. The bright region denotes higher attention scores between word and visual tokens.}}
    \label{vis}
\end{figure*}

\textbf{Image-Text Retrieval} aims retrieval an image when give a specific caption, or vice versa. 
As in Tab.\ref{res_tab1}\&\ref{res_tab2}, we achieve superior performances in all evaluation settings on both datasets, especially outperforming SOHO by 5.65\% and 4.90\% R@1 in Flickr30k-IR/-TR, 
1.71\% and 3.52\% R@1 in MSCOCO-IR/-TR 1K test set as well as 6.04\% and 7.88\% in the 5K test set. It is worthing noting that
we outperform the two-step pretraining SOTA approach UNIMO by a moderate margin, despite that they use 
additional outdomain datasets, text corpus, image collections, and adversarial training.

\textbf{Visual Entailment (VE)} predicts whether an image semantically entails the text and requires fine-grained reasoning ability in a model. 
In Tab.\ref{res_tab1}, we achieve we achieve 78.21\% accuracy in val set and 77.87\% in test set. It is worth noting that SOHO takes additional text premise as input, which leads to large improvements. 
For a fair comparison, we also implement that setting and outperform SOHO by a sizeable margin.

\textbf{NLVR$^2$} aims to determine whether a natural caption is true about a pair of photographs, which is full of semantic diversity, compositionality challenges.
We outperform SOHO, Pixel-bert, ViLT and E2E-VLP by a clear margin as in Tab.\ref{res_tab1}, and performs comparably with two-step pretraining.

\textbf{VQA} requires requires a richer multi-modal understanding to solve the free-form and open-ended questions. In Tab.\ref{res_tab1},
the results present a clear improvement compared with E2E pretraining methods while surprisingly outperform the strong two-step pretraining
methods by a slight margin.

\textbf{VCR} requires higher-order cognition and commonsense reasoning about the world. 
We achieve superior accuracy, specifically 76.70\%/78.63\%/60.53\% in three different problem setting. It is worth noting that we set up the first end-to-end benchmark for the challenging VCR task without relying on detection during inference.
Besides, we outputform VL-BERT and OSCAR by a clear margin and work comparably with VILLA, which adopts advanced adversarial training and more outdomain corpus.

Overall, our approach outperforms the previous E2E pretraining by a sizeable margin, which indicates the superiority of our object-aware E2E multi-modal representation.
In addition, we also performs better or comparably with previous state-of-the-art two-step pretrainig, like UNIMO, VILLA, Ernie-ViL, which even 
adopt more outdomain corpos, sophisticated adversarial training.

\subsection{Ablation Study \& Visualization Analysis}
In this section, we validate the effectiveness of each pretext task and provide qualitative visualization analysis.
To save experimental cost, we adopt a light-weighted ResNet-18 and 3-layer Transformer network to 
conduct the ablation study.

\paragraph{Baseline:} The baseline takes standard ITM and MLM to train
the entire model. In Tab.\ref{aba}, it still achieves decent results over various VL tasks.

\paragraph{Object-guided masked vision modeling:}
As in Tab.\ref{aba}, compared with baseline, OMVM presents a clearly consistent improvement over all downstream tasks.
It suggests that OMVM can enhance the end-to-end multi-modal representations with explicit object concepts learning.
In addition, the knowledge-guided masking strategy further helps establish cross-modal correspondence. 

To further investigate the OMVM task, we randomly mask a box region with 15\% probability rather than sampling a region based on the normalized similarity score $\alpha_{z,n}$,
denoted as RandomMVM. The other pretraining details are the same as in OMVM.
We observe a significant performance drop over all downstream tasks, especially in image-text retrieval and NLVR$^2$.
It indicates that simple RandomMVM will result in inefficient multi-modal representation learning because there is a high probability that
the selected region has no relationship with the associated description.

In addition, we also explore the similar masked feature regression task as in UNITER by randomly masking out 
the image grid features as in BERT and then requiring the Transformer to reconstruct its original features rather than the external object RoI embeddings, denoted as StandardMVM.
The results show that such StandardMVM fails to facilitate multi-modal representation learning in the E2E framework.

\begin{figure}
    \centering
    \includegraphics[width=0.85\linewidth]{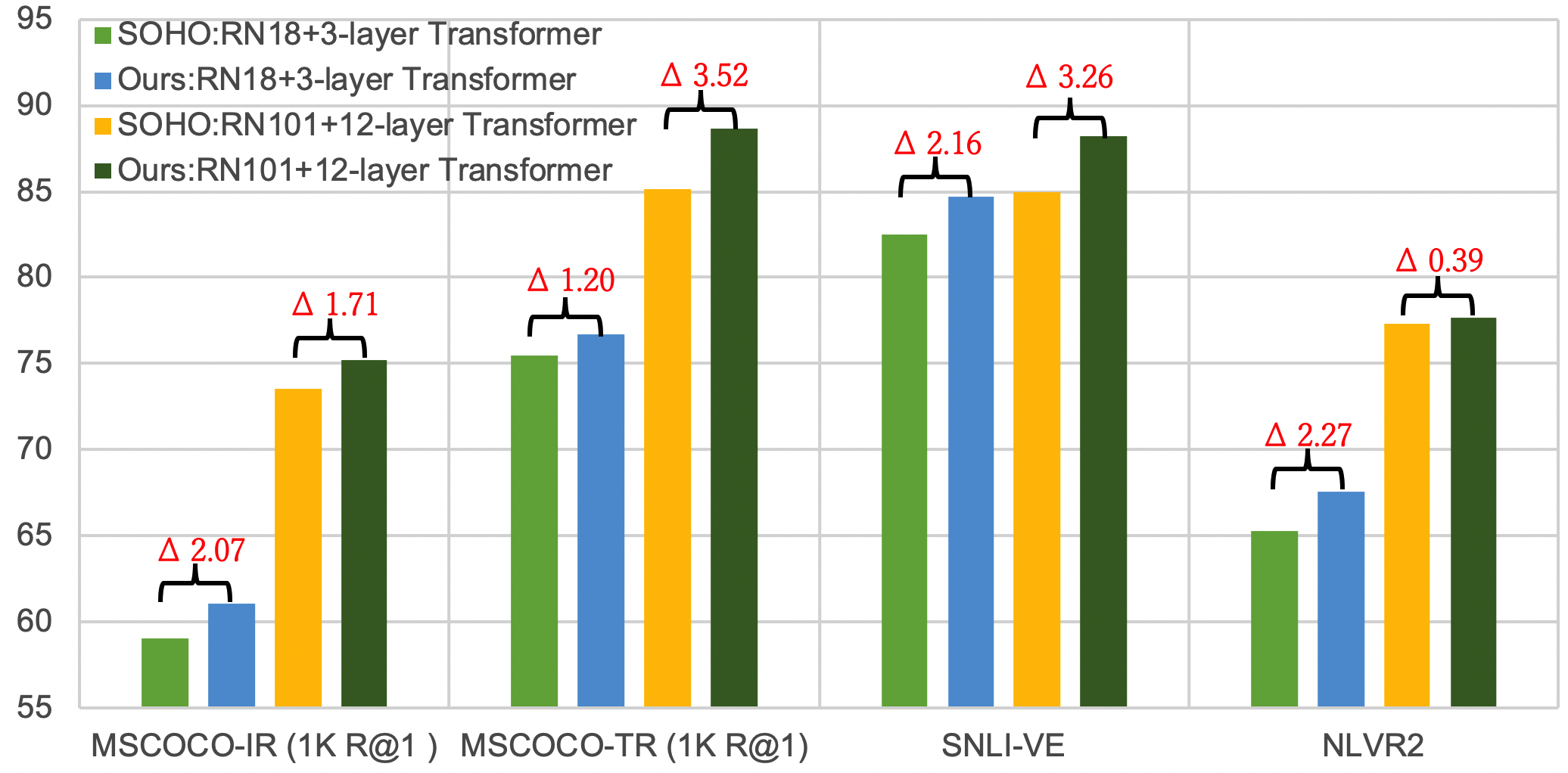}
    \caption{\small{Performance gains in different model size}}
    \label{odms}
\end{figure}

\paragraph{Phrase-region alignment:}
The OMVM above mainly focuses on instance-level knowledge distillation
by absorbing external object RoI features and semantic labels.
Different from that, PRA aims to establish positive object-phrase correspondence while suppressing the negative ones under the
guidance of similarities between noun phrases and object labels in linguistic space. As in Tab.~\ref{aba}, we significantly 
improve 0.78\% R@1 of MSCOCO-TR and 1.87\% in MSCOCO-IR. In addition,
PRA shows slight improvements for more challenging fine-grained reasoning tasks, like VE, NLVR$^2$, and VQA.  
The results indicate that PRA is beneficial to multi-modal representation learning.

\paragraph{Visualization analysis:} In Fig.\ref{vis}(a), our knowledge-guided masking strategy always masks out
the phrase-related image regions, which can facilitate multi-modal learning. On the contrary, previous works, like SOHO, VILLA ...,  mask out background regions 
or part of the object region with a high probability, which have no relationship with the corresponding description and result in
inefficient cross-modal alignment.
Fig.\ref{vis}(b) demonstrates the word-to-image attention maps. Compared to SOHO, our method can attend more accurately to image regions for the corresponding word.
Surprisingly, even the word "smiling" can locate the baby's face correctly, which suggests that our approach not only learns better noun-region alignment but also helps establish
high-order correspondence, like actions. (see Suppl. for more visualization.)

\paragraph{Influence of object detector:} We adopt the default BUTD detector in a typical 2-step pretraining 
method for a largely fair comparison. To investigate the influence of object detectors, we also conduct pretraining with objects knowledge extracted from
FRCNN-RN101 pretrained on COCO. In Tab.\ref{res_tab4}, we observe a performance drop compared with the model pretrained with BUTD, which suggests large object knowledge space
will facilitate multimodal pretraining. Besides, although with COCO detector, we still outperform SOHO by a clear margin, indicating the superiority of object knowledge in E2E pretraining
framework.

\paragraph{Contribution of each pretext task:} 
In Tab.\ref{res_tab5}, we show the individual contributions of our proposed tasks. MRC, MRFR, PRA 
pretext tasks all help facilitate multi-modal representation learning and improve the performance compared with the baseline model as a result.

\begin{table}
	\centering
	\caption{\small{Performance with different object detectors}}
	\resizebox{0.45\textwidth}{!}{
		\begin{tabular}{ccccc}
			\hline
			Models  &Detectors & Categories  &NVLR-dev &VQA-test dev      \\
			\hline
            SOHO   & -    & -     &64.62  &66.69   \\
            KD-VLP (ours) & FRCNN on COCO & 80  &65.86  &67.14   \\
            KD-VLP (ours) & BUTD & 1600  &66.71  &68.19   \\
			\hline
	 \end{tabular}}
	\label{res_tab4}
\end{table}

\begin{table}
	\centering
	\caption{\small{Individual contribution of each pretext task}}
	\resizebox{0.4\textwidth}{!}{
		\begin{tabular}{cccc}
			\hline
			Models    &Pretext Tasks   &NVLR-dev &VQA-test dev \\
            \hline
            baseline  & ITM+MLM              &62.13  &66.62   \\
            -         & ITM+MLM+MRC          &64.44  &67.27   \\
            -         & ITM+MLM+MRFR         &64.23  &67.36   \\
            -         & ITM+MLM+PRA          &63.78  &67.17   \\
            KD-VLP (ours) & ITM+MLM+MRC+MRFR+PRA &66.71  &68.19   \\
			\hline
	 \end{tabular}}
	\label{res_tab5}
\end{table}

\paragraph{Impact of object knowledge distillation in different model sizes:} 
We take SOHO as a strong baseline and compare it at different model sizes (ResNet18 + 3-layer Transformer, ResNet101 + 12-layer Transformer) 
to investigate the impact of object knowledge distillation.
Fig.\ref{odms} demonstrates the performance gains over some representative vision-language tasks.
It shows that object concepts learning always helps multi-modal representation learning no matter what model size it is. 
In VE and text-retrieval, the larger model even improves significantly than the light-weighted model and shows more capacities to learn external object semantics knowledge.

\section{Conclusion}
In this paper, we have proposed a novel self-supervised VLP method that promotes learning object-aware
multi-modal representations in an end-to-end framework. Our key idea is to perform
object knowledge distillation in both semantic and feature
space from external detectors in the pretraining stage. In particular, we develop an object-guided masked vision modeling task for distilling
external object knowledge, and a phrase-region alignment task for learning better alignment of linguistic entities and visual concepts. Compared with prior works, we achieve competitive or superior performance without relying on sophisticated object detectors during model finetuning and test in downstream tasks.

\bibliography{anthology,custom}

\begin{thebibliography}{31}
\expandafter\ifx\csname natexlab\endcsname\relax\def\natexlab#1{#1}\fi

\bibitem[{Anderson et~al.(2018)Anderson, He, Buehler, Teney, Johnson, Gould,
  and Zhang}]{anderson2018bottom}
Peter Anderson, Xiaodong He, Chris Buehler, Damien Teney, Mark Johnson, Stephen
  Gould, and Lei Zhang. 2018.
\newblock Bottom-up and top-down attention for image captioning and visual
  question answering.
\newblock In \emph{CVPR2018}.

\bibitem[{Antol et~al.(2015)Antol, Agrawal, Lu, Mitchell, Batra, Zitnick, and
  Parikh}]{antol2015vqa}
Stanislaw Antol, Aishwarya Agrawal, Jiasen Lu, Margaret Mitchell, Dhruv Batra,
  C~Lawrence Zitnick, and Devi Parikh. 2015.
\newblock Vqa: Visual question answering.
\newblock In \emph{CVPR2015}.

\bibitem[{Bowman et~al.(2015)Bowman, Angeli, Potts, and
  Manning}]{bowman2015large}
Samuel~R Bowman, Gabor Angeli, Christopher Potts, and Christopher~D Manning.
  2015.
\newblock A large annotated corpus for learning natural language inference.
\newblock \emph{arXiv preprint arXiv:1508.05326}.

\bibitem[{Carion et~al.(2020)Carion, Massa, Synnaeve, Usunier, Kirillov, and
  Zagoruyko}]{carion2020end}
Nicolas Carion, Francisco Massa, Gabriel Synnaeve, Nicolas Usunier, Alexander
  Kirillov, and Sergey Zagoruyko. 2020.
\newblock End-to-end object detection with transformers.
\newblock In \emph{ECCV2020}, pages 213--229.

\bibitem[{Chen et~al.(2020)Chen, Li, Yu, El~Kholy, Ahmed, Gan, Cheng, and
  Liu}]{chen2019uniter}
Yen-Chun Chen, Linjie Li, Licheng Yu, Ahmed El~Kholy, Faisal Ahmed, Zhe Gan,
  Yu~Cheng, and Jingjing Liu. 2020.
\newblock Uniter: Learning universal image-text representations.

\bibitem[{Devlin et~al.(2018)Devlin, Chang, Lee, and
  Toutanova}]{devlin2018bert}
Jacob Devlin, Ming-Wei Chang, Kenton Lee, and Kristina Toutanova. 2018.
\newblock Bert: Pre-training of deep bidirectional transformers for language
  understanding.
\newblock \emph{arXiv preprint arXiv:1810.04805}.

\bibitem[{Gan et~al.(2020)Gan, Chen, Li, Zhu, Cheng, and Liu}]{gan2020large}
Zhe Gan, Yen-Chun Chen, Linjie Li, Chen Zhu, Yu~Cheng, and Jingjing Liu. 2020.
\newblock Large-scale adversarial training for vision-and-language
  representation learning.
\newblock \emph{NeuIPS2020}.

\bibitem[{He et~al.(2016)He, Zhang, Ren, and Sun}]{He_2016_CVPR}
Kaiming He, Xiangyu Zhang, Shaoqing Ren, and Jian Sun. 2016.
\newblock Deep residual learning for image recognition.
\newblock In \emph{CVPR2016}.

\bibitem[{Huang et~al.(2021)Huang, Zeng, Huang, Liu, Fu, and
  Fu}]{huang2021seeing}
Zhicheng Huang, Zhaoyang Zeng, Yupan Huang, Bei Liu, Dongmei Fu, and Jianlong
  Fu. 2021.
\newblock Seeing out of the box: End-to-end pre-training for vision-language
  representation learning.
\newblock \emph{CVPR2021}.

\bibitem[{Huang et~al.(2020)Huang, Zeng, Liu, Fu, and Fu}]{huang2020pixel}
Zhicheng Huang, Zhaoyang Zeng, Bei Liu, Dongmei Fu, and Jianlong Fu. 2020.
\newblock Pixel-bert: Aligning image pixels with text by deep multi-modal
  transformers.
\newblock \emph{arXiv preprint arXiv:2004.00849}.

\bibitem[{Kim et~al.(2021)Kim, Son, and Kim}]{kim2021vilt}
Wonjae Kim, Bokyung Son, and Ildoo Kim. 2021.
\newblock Vilt: Vision-and-language transformer without convolution or region
  supervision.
\newblock \emph{ICML2021}.

\bibitem[{Krishna et~al.(2016)Krishna, Zhu, Groth, Johnson, Hata, Kravitz,
  Chen, Kalantidis, Li, Shamma et~al.}]{krishna2016visual}
Ranjay Krishna, Yuke Zhu, Oliver Groth, Justin Johnson, Kenji Hata, Joshua
  Kravitz, Stephanie Chen, Yannis Kalantidis, Li-Jia Li, David~A Shamma, et~al.
  2016.
\newblock Visual genome: Connecting language and vision using crowdsourced
  dense image annotations.
\newblock \emph{arXiv preprint arXiv:1602.07332}.

\bibitem[{Lee et~al.(2018)Lee, Chen, Hua, Hu, and He}]{lee2018stacked}
Kuang-Huei Lee, Xi~Chen, Gang Hua, Houdong Hu, and Xiaodong He. 2018.
\newblock Stacked cross attention for image-text matching.
\newblock In \emph{ECCV2018}.

\bibitem[{Li et~al.(2020{\natexlab{a}})Li, Duan, Fang, Gong, and
  Jiang}]{li2020unicoder}
Gen Li, Nan Duan, Yuejian Fang, Ming Gong, and Daxin Jiang. 2020{\natexlab{a}}.
\newblock Unicoder-vl: A universal encoder for vision and language by
  cross-modal pre-training.
\newblock In \emph{AAAI2020}.

\bibitem[{Li et~al.(2019)Li, Yatskar, Yin, Hsieh, and Chang}]{li2019visualbert}
Liunian~Harold Li, Mark Yatskar, Da~Yin, Cho-Jui Hsieh, and Kai-Wei Chang.
  2019.
\newblock Visualbert: A simple and performant baseline for vision and language.
\newblock \emph{arXiv preprint arXiv:1908.03557}.

\bibitem[{Li et~al.(2021)Li, Gao, Niu, Xiao, Liu, Liu, Wu, and
  Wang}]{li2020unimo}
Wei Li, Can Gao, Guocheng Niu, Xinyan Xiao, Hao Liu, Jiachen Liu, Hua Wu, and
  Haifeng Wang. 2021.
\newblock Unimo: Towards unified-modal understanding and generation via
  cross-modal contrastive learning.
\newblock \emph{ACL2021}.

\bibitem[{Li et~al.(2020{\natexlab{b}})Li, Yin, Li, Zhang, Hu, Zhang, Wang, Hu,
  Dong, Wei et~al.}]{li2020oscar}
Xiujun Li, Xi~Yin, Chunyuan Li, Pengchuan Zhang, Xiaowei Hu, Lei Zhang, Lijuan
  Wang, Houdong Hu, Li~Dong, Furu Wei, et~al. 2020{\natexlab{b}}.
\newblock Oscar: Object-semantics aligned pre-training for vision-language
  tasks.
\newblock In \emph{ECCV2020}.

\bibitem[{Lin et~al.(2014)Lin, Maire, Belongie, Hays, Perona, Ramanan,
  Doll{\'a}r, and Zitnick}]{lin2014microsoft}
Tsung-Yi Lin, Michael Maire, Serge Belongie, James Hays, Pietro Perona, Deva
  Ramanan, Piotr Doll{\'a}r, and C~Lawrence Zitnick. 2014.
\newblock Microsoft coco: Common objects in context.
\newblock In \emph{ECCV2014}.

\bibitem[{Liu et~al.(2021)Liu, Wan, Ma, and He}]{Liu_2021_CVPR}
Yongfei Liu, Bo~Wan, Lin Ma, and Xuming He. 2021.
\newblock Relation-aware instance refinement for weakly supervised visual
  grounding.
\newblock In \emph{CVPR2021}.

\bibitem[{Liu et~al.(2020)Liu, Wan, Zhu, and He}]{liu2020learning}
Yongfei Liu, Bo~Wan, Xiaodan Zhu, and Xuming He. 2020.
\newblock Learning cross-modal context graph for visual grounding.
\newblock In \emph{AAAI}.

\bibitem[{Lu et~al.(2019)Lu, Batra, Parikh, and Lee}]{lu2019vilbert}
Jiasen Lu, Dhruv Batra, Devi Parikh, and Stefan Lee. 2019.
\newblock Vilbert: Pretraining task-agnostic visiolinguistic representations
  for vision-and-language tasks.
\newblock \emph{NeuIPS2019}.

\bibitem[{Park et~al.(2020)Park, Bhagavatula, Mottaghi, Farhadi, and
  Choi}]{park2020visualcomet}
Jae~Sung Park, Chandra Bhagavatula, Roozbeh Mottaghi, Ali Farhadi, and Yejin
  Choi. 2020.
\newblock Visualcomet: Reasoning about the dynamic context of a still image.
\newblock In \emph{ECCV2020}.

\bibitem[{Plummer et~al.(2015)Plummer, Wang, Cervantes, Caicedo, Hockenmaier,
  and Lazebnik}]{plummer2015flickr30k}
Bryan~A Plummer, Liwei Wang, Chris~M Cervantes, Juan~C Caicedo, Julia
  Hockenmaier, and Svetlana Lazebnik. 2015.
\newblock Flickr30k entities: Collecting region-to-phrase correspondences for
  richer image-to-sentence models.
\newblock In \emph{ICCV2015}.

\bibitem[{Su et~al.(2020)Su, Zhu, Cao, Li, Lu, Wei, and Dai}]{su2019vl}
Weijie Su, Xizhou Zhu, Yue Cao, Bin Li, Lewei Lu, Furu Wei, and Jifeng Dai.
  2020.
\newblock Vl-bert: Pre-training of generic visual-linguistic representations.
\newblock \emph{ICLR2020}.

\bibitem[{Tan and Bansal(2019)}]{tan2019lxmert}
Hao Tan and Mohit Bansal. 2019.
\newblock Lxmert: Learning cross-modality encoder representations from
  transformers.
\newblock \emph{arXiv preprint arXiv:1908.07490}.

\bibitem[{Wu et~al.(2016)Wu, Schuster, Chen, Le, Norouzi, Macherey, Krikun,
  Cao, Gao, Macherey et~al.}]{wu2016google}
Yonghui Wu, Mike Schuster, Zhifeng Chen, Quoc~V Le, Mohammad Norouzi, Wolfgang
  Macherey, Maxim Krikun, Yuan Cao, Qin Gao, Klaus Macherey, et~al. 2016.
\newblock Google's neural machine translation system: Bridging the gap between
  human and machine translation.
\newblock \emph{arXiv preprint arXiv:1609.08144}.

\bibitem[{Xie et~al.(2019)Xie, Lai, Doran, and Kadav}]{xie2019visual}
Ning Xie, Farley Lai, Derek Doran, and Asim Kadav. 2019.
\newblock Visual entailment: A novel task for fine-grained image understanding.
\newblock \emph{arXiv preprint arXiv:1901.06706}.

\bibitem[{Xu et~al.(2021)Xu, Yan, Li, Bi, Huang, Xiao, and Huang}]{xu2021e2e}
Haiyang Xu, Ming Yan, Chenliang Li, Bin Bi, Songfang Huang, Wenming Xiao, and
  Fei Huang. 2021.
\newblock E2e-vlp: End-to-end vision-language pre-training enhanced by visual
  learning.
\newblock \emph{ACL2021}.

\bibitem[{Yu et~al.(2020)Yu, Tang, Yin, Sun, Tian, Wu, and Wang}]{yu2020ernie}
Fei Yu, Jiji Tang, Weichong Yin, Yu~Sun, Hao Tian, Hua Wu, and Haifeng Wang.
  2020.
\newblock Ernie-vil: Knowledge enhanced vision-language representations through
  scene graph.
\newblock \emph{arXiv preprint arXiv:2006.16934}.

\bibitem[{Zhang et~al.(2021)Zhang, Li, Hu, Yang, Zhang, Wang, Choi, and
  Gao}]{Zhang_2021_CVPR}
Pengchuan Zhang, Xiujun Li, Xiaowei Hu, Jianwei Yang, Lei Zhang, Lijuan Wang,
  Yejin Choi, and Jianfeng Gao. 2021.
\newblock Vinvl: Revisiting visual representations in vision-language models.
\newblock In \emph{CVPR2021}.

\bibitem[{Zhou et~al.(2021)Zhou, Palangi, Zhang, Hu, Corso, and
  Gao}]{zhou2020unified}
Luowei Zhou, Hamid Palangi, Lei Zhang, Houdong Hu, Jason Corso, and Jianfeng
  Gao. 2021.
\newblock Unified vision-language pre-training for image captioning and vqa.
\newblock In \emph{AAAI2020}.

\end{thebibliography}
\bibliographystyle{acl_natbib}

\appendix

\section*{Appendix}
In this supplementary material, we firstly discuss the limitations of work, then give the detailed dataset statistics of pretraining 
and each downstream task, and depict more advanced implementation details of the pretraining.
In addition, we also demonstrate how to generate a binary mask for each object proposal, 
followed by detailed experimental setups and finetuning strategies of downstream tasks. Besides, we also discuss
the influence of image size during pretraining stge.
Finally, we provide more qualitative visualization results for better understanding.
\vspace{-1mm}
\section{Limitations}
In this paper, we only pretrain our proposed KD-VLP framework on indomain datasets, including MSCOCO and Visual Genome caption datasets.
In the future, we need to scale up our model pretrained on more noisy web image-text pairs to make it to learn more general knowledge.

\section{Experiments}
\subsection{Dataset Statistics}
As shown in Tab.\ref{label} we summarize the dataset statistics of pretraining and each downstream task, 
including the number of image-text pairs and number of images for each dataset split.
It is worth mentioning that we select the MSCOCO \& Visual Genome image-text data as our pretraining datasets since
they are typical indomain datasets for many downstream tasks and are widely adopted by prior works.

\begin{table}[ht]
    \caption{\small{Dataset statistics of pretraining and downstream tasks. The number in brackets indicates the number of images}}
    \vspace{-2mm}
	\centering{
        \resizebox{0.48\textwidth}{!}{
		\begin{tabular}{c|c|c|c|c}
			\hline
			task &data sources & training & val & test\\
            \hline
            \multirow{2}{*}{pretraining} & MSCOCO & \multirow{2}{*}{5.1M(207K)}&\multirow{2}{*}{131K(7.1K)}&\multirow{2}{*}{-}\\
            &Visual Genome&&&\\
            \hline
            \multirow{2}{*}{VCR} & MovieClips & \multirow{2}{*}{213K(80K)} & \multirow{2}{*}{26.5K(9.9K)}&\multirow{2}{*}{25.2K(9.5K)}\\
            &LSMDC&&&\\
            \hline 
            Image-text & Flickr30k &145K(29K)&5K(1K)&5K(1K) \\
            \cline{2-5}
            Matching &MSCOCO &567K(113.2K)&25K(5K)&25K(5K)\\
            \hline 
            Visual &Flickr30k&\multirow{2}{*}{52.9K(29.7K)}&\multirow{2}{*}{17.8K(1K)}&\multirow{2}{*}{17.9K(1K)}\\
            Entailemnt & SNLI &&&\\
            \hline
            \multirow{2}{*}{VQA} & MSCOCO & \multirow{2}{*}{443.8(82.8K)} & \multirow{2}{*}{214.4K(40.5K)} &\multirow{2}{*}{447.8K(81.4K)}\\
             &Abstract Scenes &&&\\
            \hline
            NLVR$^2$& Flickr30k &529.5K(29.8K)&17.9K(1K)&17.9(1K) \\
            \hline
	    \end{tabular}
        }}
    \label{label}
    \vspace{-2mm}
\end{table}
\subsection{More Pretraining Details}
In pretraining stage, we also adopt gradient accumulation\footnote{https://nvidia.github.io/apex/advanced.html} and gradient checkpointing\footnote{https://pytorch.org/docs/stable/checkpoint.html} techniques to further reduce 
the GPU memory footprint and increase the batch-size. In our experiments, the gradient accumulation step size is set as 4.

\begin{figure}
    \centering
    \includegraphics[width=0.85\linewidth]{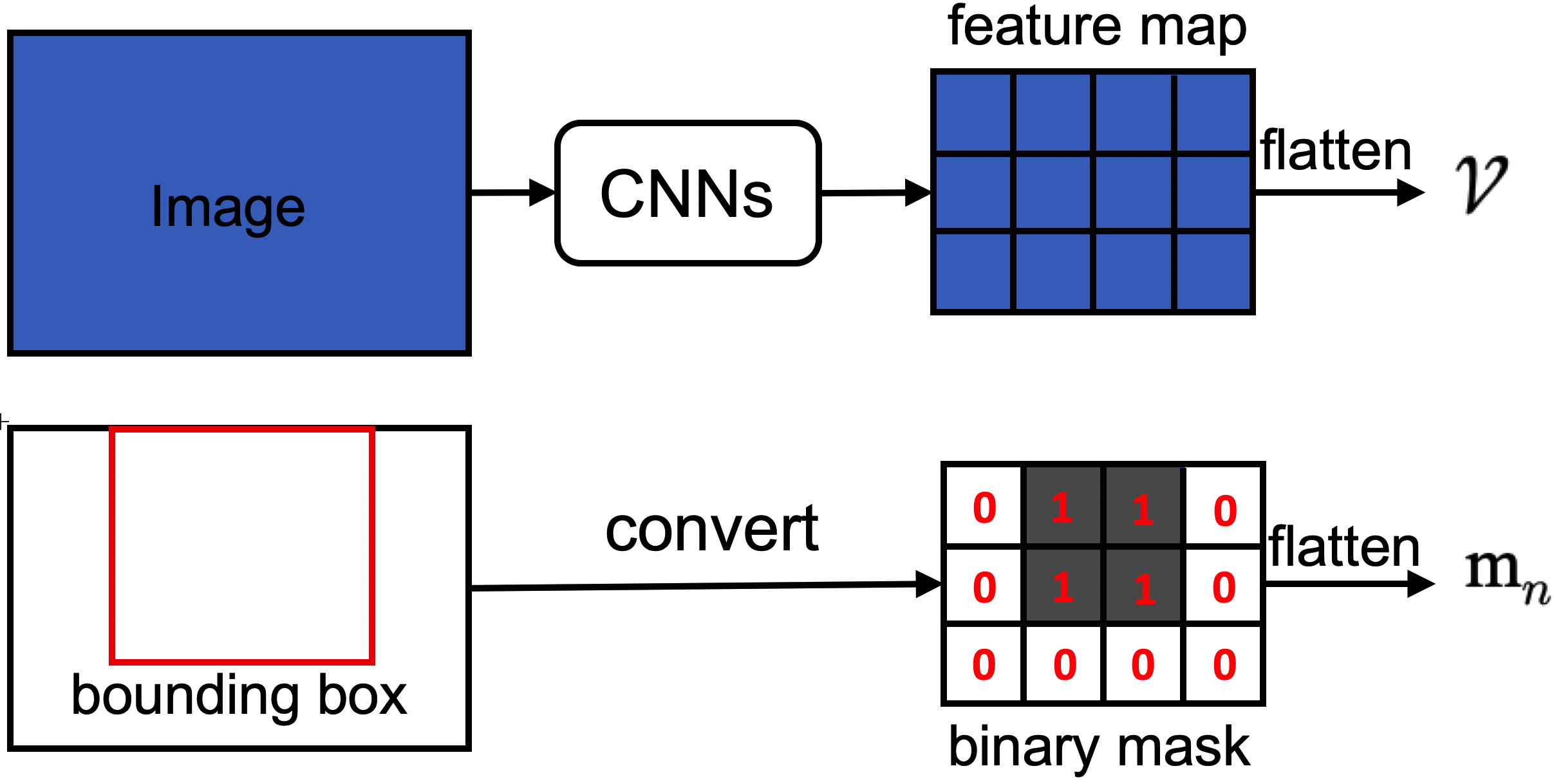}
    \caption{\small{Illustration of generating binary mask for each proposal.}}
    \label{bin}
    \vspace{-4mm}
\end{figure}

\begin{figure*}
    \begin{subfigure}{1.0\textwidth}
    \centering
    \includegraphics[width=0.85\linewidth]{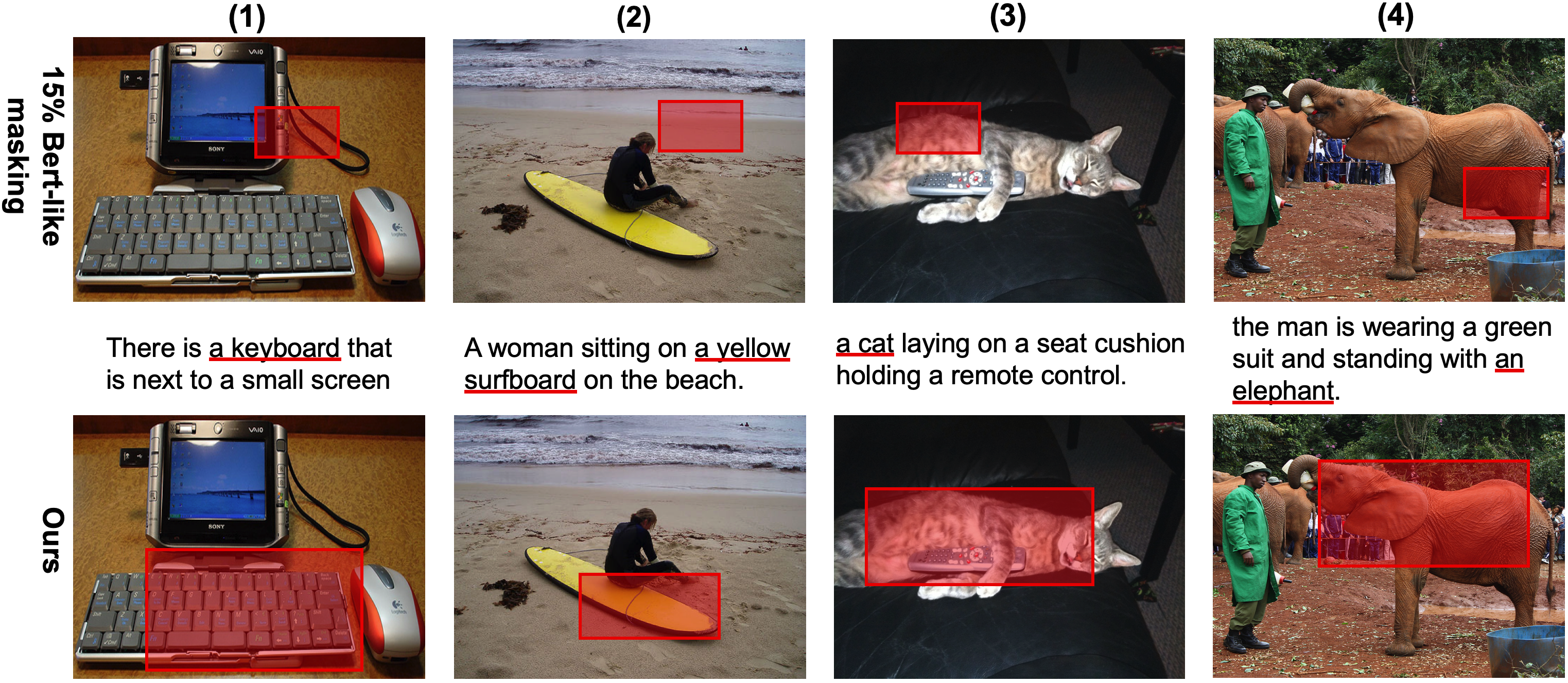}
    \caption{\small{Knowledge-guided masking strategy. \textcolor{red}{Red mask} denotes the masked region in an image}}
    \label{fig:kgms}
    \end{subfigure}

    \begin{subfigure}{1.0\textwidth}
    \centering
    \includegraphics[width=0.85\linewidth]{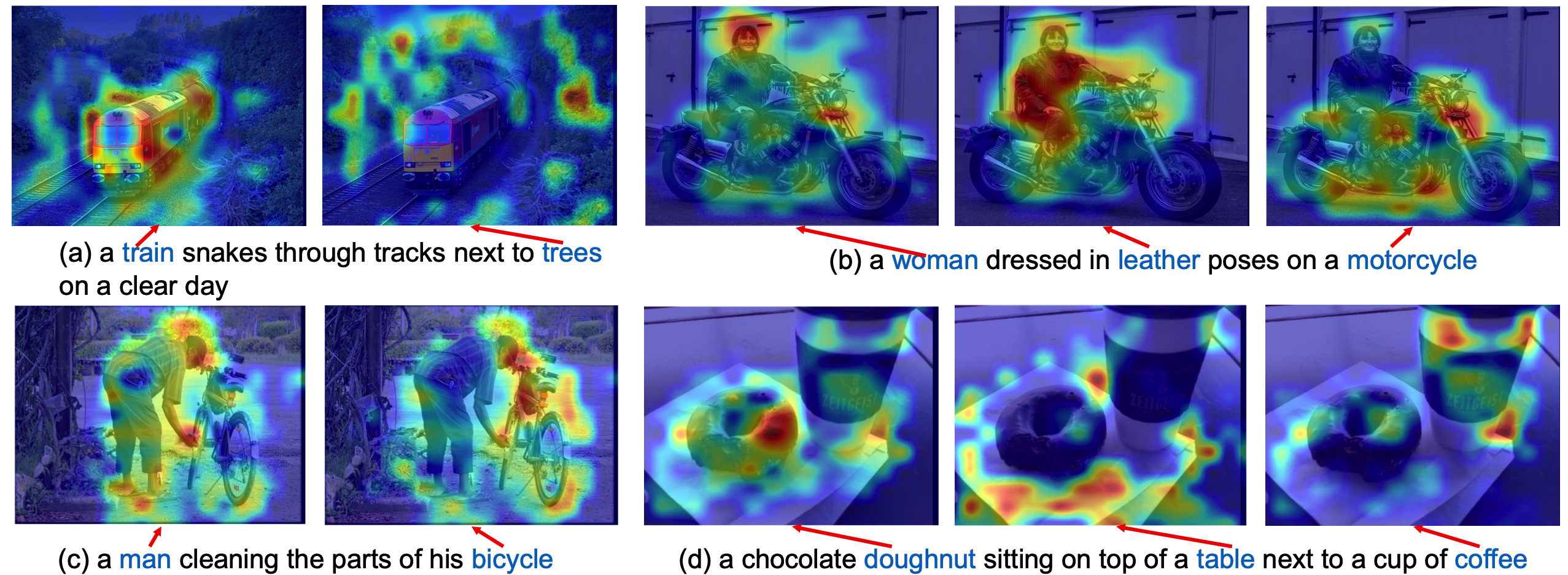}
    \caption{\small{Text-to-image attention maps. The bright region denotes higher attention scores between word tokens and image regions.}}
    \label{fig:tima}    
    \end{subfigure}
\end{figure*}

\subsection{Binary mask for each proposal}
As shown in Fig.\ref{bin}, we generate a binary mask of the same size of feature map for each proposal where
locations within the bounding box fill 1 and others fill 0.
\subsection{Detailed experiment setup for each downstream task}
\paragraph{Image-Text Retrieval:} The image-text retrieval typically includes two sub-tasks: image-retrieval (\textbf{IR}) aims 
to retrieval an image when given a specific caption and text-retrieval (\textbf{TR}) is on the contrary.
We perform experiments on both Flickr30k~\cite{plummer2015flickr30k} and MSCOCO dataset. 
As in UNITER, we construct a mini-batch for each GPU of a matched image-text pair, \textit{t}-1 negative images, and \textit{t}-1 negative texts where
$t$ is set as 32. Besides, we take a fully-connected network on top of $\mathbf{h}_{{\rm cls}}$ and adopt the binary cross-entropy loss as supervision signal.
The finetuning iterations are up to 10K by following linear decay scheduling with initial lr 7e-5 for Transformer, 1e-4 for CNNs. Top-K (R@K, $K\in\{1, 5, 10\}$) recall is the evaluation metric.

\paragraph{Visual Entailment (VE):} VE task aims to predict whether an image semantically entails the text and requires fine-grained reasoning ability in a model.
VE dataset is built upon SNLI~\cite{bowman2015large} and Flickr30k. 
Each image-text pair is assigned with one of three classes: \textit{entailment, neutral, contradiction}.
As in UNITER, we formulate it as 3-way classification problem based on $\mathbf{h}_{{\rm \mathbf{cls}}}$. 
The batch size is 32 per GPU while other finetuning strategies are the same.

\paragraph{Natural Language Visual Reasoning (NLVR$^2$): }
NLVR$^2$ aims to determine whether a natural caption is true about a pair of photographs, which is full of semantic diversity, compositionality challenges.
We follow UNITER to construct two image-text pairs for each sample and concatenate their $\mathbf{h}_{\mathbf{cls}}$ features to infer \textit{true} or \textit{false}.
All finetuning strategies are the same as before except for a batch size 12 per GPU.

\paragraph{Visual Question Answering (VQA):} VQA requires a richer multi-modal understanding to solve the free-form and open-ended questions.
VQA dataset contains 204K images from MSCOCO, 614K free-from nature language question and around 6M answers.
It is typically formulated as a 3192-way classification problem and supervised by binary cross-entropy loss as in UNITER.
The batch size here is 32 per GPU while other finetuning strategies are kept the same.

\paragraph{Visual Commonsense Reasoning (VCR):} Given a question for an image, VCR needs to 1.) correctly answer (Q$\rightarrow$A); 2.) provide
a rationale justifying its answer (QA$\rightarrow$R); 3.) reason both of them (Q$\rightarrow$AR), which requires higher-order cognition and 
commonsense reasoning about the world. Following UNITER, we introduce a second-stage pretraining over the VCR dataset due to severe difference in dataset distribution compared to 
indomain image-text corpus. In addition, we also utilize a similar person grounding~\cite{park2020visualcomet} pretext task to tightly
align the person tags in text and their visual locations.
During finetuning stage, we concatenate each question along with each possible answer to form four kinds of 
text inputs, and feed each of them into Transformer network with corresponding image embeddings. Finally, 
a binary cross-entropy loss is adopted to supervise each pair.
Since VCR questions explicitly reference objects at specific locations, we implement coreferencing between text and image 
by replacing referenced entities in the questions with their corresponding box locations. In the second stage pretraining for VCR, 
we reduce the learning rate to a constant 5e-05 and trained for an additional 9K steps. 
Due to longer sequence lengths in the VCR dataset, a training batch-size of 224 is used. We also use a step size of 2 for gradient accumulation. 
After pretraining, we finetuned on the VCR task for 10K steps with a learning rate of 1e-04 for both the Transformer and the CNNs. 
Linear warmup of the learning rate is applied for 1000 steps, followed by a linear decay ending at a total of 10K steps. 

\subsection{Influence of image size}
We adopt larger image size mainly for fair comparisons with most 2-step pretraining methods, 
PixelBert and E2E-VLP as all of them use the size (800, 1333).
To investigate this, we pretrain our method with size (600,1000) and report the results in Tab.\ref{res_tab8}. 
We can see that our method has a mild performance drop, but still outperforms SOHO by a decent margin.
\begin{table}
	\centering
	\caption{\small{Performance comparison of different image size}}
    \vspace{-3mm}
	\resizebox{0.4\textwidth}{!}{
		\begin{tabular}{cccc}
			\hline
			Models    &Image Size   &NVLR-dev &VQA-test dev      \\
			\hline            
    
            SOHO  & (600,1000)          &64.62  &66.69   \\
            KD-VLP(ours) & (600,1000)   &66.52  &68.04   \\
            KD-VLP(ours) & (800,1333)   &66.71  &68.19   \\
			\hline
	 \end{tabular}}
	\label{res_tab8}
\end{table}

\subsection{More Visualizations}
As in Fig.\ref{fig:kgms}, we observe that our knowledge-guided masking strategy masks out the image regions, which are highly related to the corresponding sentences.
This design can force Transformer to infer object features and semantic labels based on the surrounding visual context and its language descriptions.
On the contrary, SOHO randomly masks out either background regions (Fig.\ref{fig:kgms}(1) \& Fig.\ref{fig:kgms}(2)) or local object parts (Fig.\ref{fig:kgms}(3) \& Fig.\ref{fig:kgms}(4)), which are not related to the corresponding sentences with a high probability and result in 
inefficient multi-modal representation learning.

As shown in Fig.\ref{fig:tima}, it shows that our object-aware end-to-end multi-modal representations can accurately
establish the correspondence between word tokens and visual tokens, which demonstrates the superiority of our approach.

\end{document}